\documentclass[lettersize,journal]{IEEEtran}
\usepackage{amsmath,amsfonts}
\usepackage{algorithmic}
\usepackage{algorithm}
\usepackage{array}
\usepackage[caption=false,font=normalsize,labelfont=sf,textfont=sf]{subfig}
\usepackage{textcomp}
\usepackage{stfloats}
\usepackage{url}
\usepackage{verbatim}
\usepackage{graphicx}
\usepackage{cite}
\usepackage{xcolor}
\usepackage{booktabs}
\usepackage{comment}
\usepackage{makecell}
\usepackage{hyperref}
\usepackage{soul}

\begin{document}

\title{Multispectral Blind Image Super-Resolution \\ for Standing Dead Tree Segmentation}

\author{Mete Ahishali, Anis Ur Rahman, Einari Heinaro, Aysen Degerli, and Samuli Junttila
\thanks{Mete Ahishali is with the Department of Forest Sciences, University of Helsinki, Finland (email: \textit{mete.ahishali@helsinki.fi}).}
\thanks{Anis Ur Rahman is with CSC – IT Center for Science Ltd., Espoo, Finland (email: \textit{anis.rahman@csc.fi}).}
\thanks{Einari Heinaro is with KOKO Forest Ltd. (email: \textit{einari.heinaro@kokoforest.com}).}
\thanks{Aysen Degerli is with VTT Technical Research Centre of Finland, Finland (email: \textit{aysen.degerli@vtt.fi}).}
\thanks{Samuli Junttila is with the Department of Forest Sciences, University of Helsinki, Finland, and the School of Forest Sciences, University of Eastern Finland, Finland (email: \textit{samuli.junttila@helsinki.fi}; \textit{samuli.junttila@uef.fi}).}
}



\maketitle

\begin{abstract}
Mapping standing dead trees is crucial for acquiring information on the effects of climate change on forests and forest biodiversity. However, leveraging high-quality aerial imagery for dead tree segmentation poses challenges due to limitations in sensor availability and the scarcity of annotated data. In this study, we propose a generic blind super-resolution framework that incorporates Attention-Guided Domain Adaptation Networks (ADA-Nets) to learn the mapping from low-resolution to high-resolution multispectral image domains. Our approach operates solely on unpaired samples, mimicking real-world conditions, i.e., low-resolution images are not synthetically obtained by downsampling the high-resolution images. Moreover, the proposed method serves as a general-purpose restorer addressing several image degradation types, including saturation, noise, and low contrast that typically occur in low-resolution images acquired by low-end sensors. To the best of our knowledge, this is the first study to perform real-world and generic super-resolution for multispectral data in the scope of standing dead tree segmentation. Experimental evaluations demonstrate segmentation performances of $54\%$ and $64\%$ in Dice scores. Notably, the first result is obtained without using any high-resolution annotations; the segmentation network is trained on super-resolved low-resolution images, while evaluation is performed on the high-resolution data. We publicly share the multispectral dataset with manually annotated labels \footnote{The multispectral aerial imagery dataset with annotations is available at \href{https://www.kaggle.com/datasets/meteahishali/aerial-imagery-for-dead-tree-segmentation-poland}{https://www.kaggle.com/datasets/meteahishali/aerial-imagery-for-dead-tree-segmentation-poland}.}.
\end{abstract}

\begin{IEEEkeywords}
Aerial imagery, dead tree segmentation, generative adversarial networks, image-to-image translation, super-resolution.
\end{IEEEkeywords}

\section{Introduction}
\label{sec:introduction}
\IEEEPARstart{S}{uper}-resolution (SR) is defined as reconstructing a high-resolution image from a given low-resolution input, aiming to improve overall spatial and visual quality. The SR task plays a pivotal role in many remote sensing applications, including satellite imaging and Earth observation \cite{8884136}, where the high-resolution data are costly or infeasible due to the sensor limitations. Super-resolution methods are typically divided into two groups \cite{anwar2020deep}: multi-image \cite{8884136, Arefin_2020_CVPR_Workshops} and single-image \cite{8578438, CHEN2022124} super-resolution. In the first group of approaches, multiple low-resolution images capturing similar regions-of-interest are utilized to generate a single high-quality image. In this work, we focus on the latter group of approaches, specifically those concerning single-image super-resolution, due to their cost-effectiveness and wider applicability. Among them, many approaches \cite{lu2022transformer, yang2014single, xiao2024frequency, 11329206, 11435474} have been proposed to exploit the relation between the low- and high-resolution images by training neural networks and approximating the mapping function in an end-to-end manner using paired training samples. 

A wide variety of learning-based super-resolution methods have been proposed in the literature. Efficient transformer architectures for natural image-super-resolution are introduced in \cite{lu2022transformer} demonstrating the capability of capturing long-range dependencies. In \cite{yang2014single}, a benchmark evaluation method is proposed  to systematically compare various super-resolution techniques. Next, the study in \cite{xiao2024frequency} has investigated the frequency-assisted super-resolution for remote sensing images, where the frequency-domain information is utilized to enlarge the effective receptive field of neural networks approach. This is especially important considering the large spatial dimensions of remote sensing images. The study in \cite{11329206} and \cite{11435474} have focused on super-resolution of hyperspectral images, exploring scenarios with fixed and varying scaling factors. Specifically, authors in \cite{11329206} argue the necessity of dedicated upsampling layers when processing high-dimensional hyperspectral data, whereas the study in \cite{11435474} highlights the limited attention paid to super-resolved boundaries in the existing approaches.

\IEEEpubidadjcol

Despite their methodological differences, all of these methods \cite{lu2022transformer, yang2014single, xiao2024frequency, 11329206, 11435474} rely on low-resolution images that are synthetically produced by downsampling high-resolution counterparts. This common practice raises concerns about its ability to replicate the characteristics inherent to real-world low-resolution imagery for several reasons. First, simulated low-resolution images often lack the same spatial resolution characteristics as real low-resolution images. Secondly, since low-resolution images are captured by inexpensive sensors, the low-quality domain, i.e., low-resolution images, tends to contain additional degradation such as low contrast, increased noise, and saturation artifacts. These properties are underrepresented because of artificially downsampling images, which mainly addresses only the spatial resolution aspect. The proposed approach in \cite{cai2019toward} is a pioneering effort towards a real-world super-resolution technique. The authors have constructed a dataset comprising low- and high-resolution image pairs acquired by the same camera, but using different focal lengths. While such a paired data collection procedure is more realistic than using simulated data, their approach still fails to address the aforementioned key limitation, i.e., low- and high-resolution images are acquired with different sensors in the real-world image super-resolution problem, resulting in variations in overall image quality.

Vision Transformers (ViT) \cite{dosovitskiy2021an} and Convolutional Neural Networks (CNNs) represent the current state-of-the-art models in computer vision. Expectedly, there has been a growing interest in utilizing these models in tree mortality mapping applications. Tree mortality mapping is essential for many environmental applications, including forest health monitoring, exploring climate change impacts, biodiversity assessments, and evaluating different ecological disturbances such as deforestation and wildfires \cite{deng2024individual, international2025towards}. For this purpose, many studies \cite{rahman2025dual, chiang2020deep, SANIMOHAMMED2022100024, cheng2024scattered, duarte2021machine} have proposed segmenting standing dead trees using remote sensing. Particularly, utilizing aerial imagery is a cost-efficient and accessible alternative to LiDARs, considering its wider spatial coverage in forest monitoring applications. Various studies leverage aerial imagery and employ methods based on ViT \cite{rahman2025dual}, CNNs \cite{chiang2020deep, SANIMOHAMMED2022100024, cheng2024scattered}, and traditional machine learning with Support Vector Machines and random forests \cite{duarte2021machine}. Notably, the proposed approach in \cite{rahman2025dual} utilized a ViT-based U-Net segmentation framework with a dual-task learning strategy. The dual-task learning, TreeMort-1T-UNet and TreeMort-3T-UNet models \cite{rahman2025dual}, involves detecting dead tree centroids and segmenting canopy shapes for finer details. However, transformer models have a common key limitation, which is the need for large training data \cite{bai2021transformers}. This is particularly important in forest health monitoring applications, where annotated datasets are scarce. Therefore, the real-world deployment of dead tree detection frameworks is often impractical because of a resource-intensive labeling procedure, which is time-consuming and relies on human expertise with specialized domain knowledge, sometimes even requiring field reference.

In this study, we propose interpreting the super-resolution task as an image-to-image translation objective. The source domain consists of average-quality images affected mainly by low-resolution and poor-quality ones that exhibit additional degradations such as poor contrast and high saturation. The proposed approach in this work is based on our recent study Attention-Guided Domain Adaptation Network (ADA-Net) with contrastive learning \cite{ahishali2025ada}, where we adapted ADA-Net to learn the mapping from low-resolution and mixed quality source domain to high-quality target domain. The high-quality domain images often feature finer details, captured by more advanced high-resolution sensors. We show that the proposed image-to-image translation framework for super-resolution improves the segmentation of standing dead trees in our experimental evaluations. Moreover, rich spectral information provides enhanced interpretation of the target with improved separability between vegetation and non-natural features, enabled by a 4-band multispectral super-resolution approach. The contributions of this study are summarized as follows,
\begin{itemize}
    \item We publicly share the annotations of standing dead trees in high-resolution multispectral aerial imagery labeled by forest health experts.
    \item To the best of our knowledge, this work proposes the first generic super-resolution approach for multispectral data within the scope of tree mortality mapping.
    \item The proposed generic image restoration approach operates in a fully blind manner, without any prior assumptions about the nature of image degradation, such as its intensity, type, or downsampling factor.
\end{itemize}

The rest of the paper is organized as follows: Section \ref{sec:methodology} introduces the methodology involving image-to-image translation for super-resolution and ADA-Nets. Next, an extensive set of experimental evaluations is presented in Section \ref{sec:experimental_evaluation}, which is followed by Section \ref{sec:conclusion}, concluding this study with future work directions.

\section{Methodology}
\label{sec:methodology}

\begin{figure*}[!h]
    \centering
    \includegraphics[width=1\linewidth]{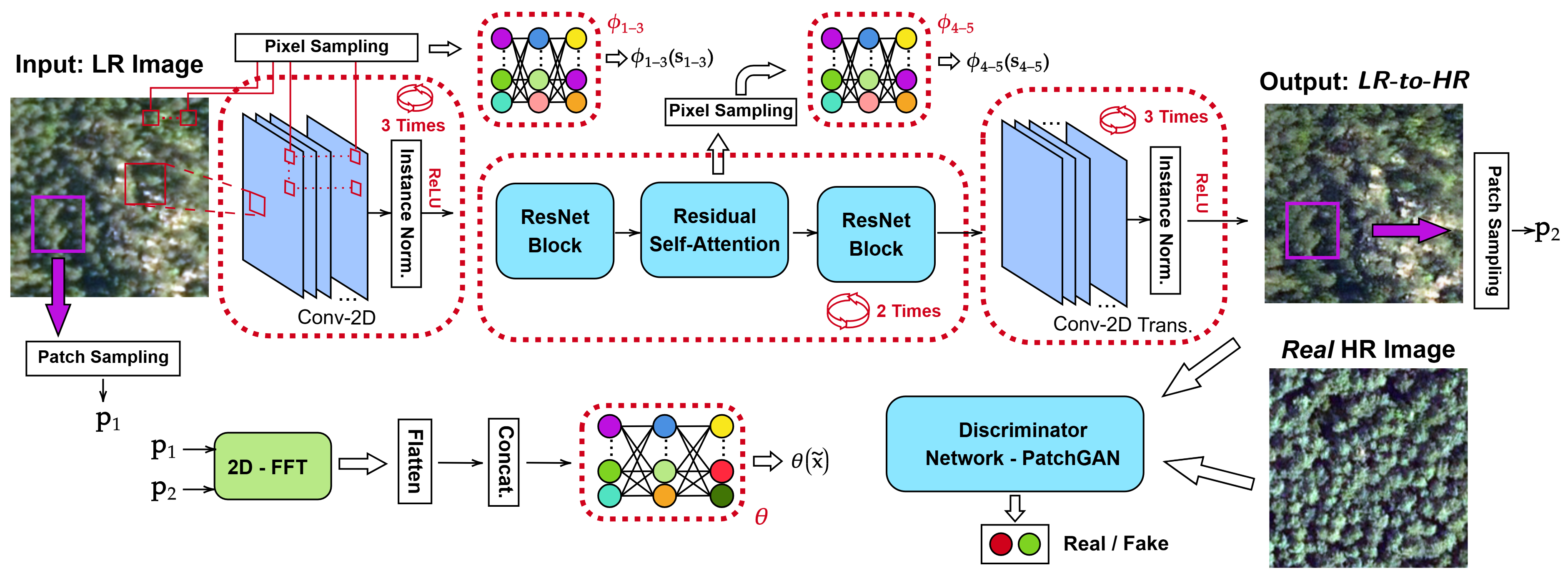}
    \caption{Blind super-resolution framework with Attention-Guided Domain Adaptation Network (ADA-Net). The generator learns low-resolution (LR) to high-resolution (HR) mapping using unpaired training samples, where the discriminator tries to distinguish between the synthetic HR (super-resolved) images and real-domain images. For illustration purposes, we use RGB representation, whereas the image domains comprise 4-band multispectral data.}
\label{fig:framework}
\end{figure*}

Let $\mathbf{I}_{l, i} \in I_L$, $\mathbf{I}_{h, j} \in I_H$ be \textit{unpaired} training samples, i.e., $i \neq j$, where $I_L, I_H \in \mathbb{R}^{P \times N \times C}$ are low- and high-resolution image domains, respectively. To streamline notation, we omit the indices $i,j$ throughout the remainder of this paper. The ADA-Net approach for super-resolution is presented in Fig. \ref{fig:framework}, which consists of a generator and a discriminator network. The generator provides the following mapping: $G: I_L \rightarrow I_H$. While the discriminator architecture is the same as the conventional PatchGAN classifier proposed in the \textit{pix2pix} model \cite{isola2017image}, the generator network consists of four residual blocks \cite{7780459} and residual self-attention layers \cite{ahishali2025ada}. Let a feature tensor of an intermediate layer be $\mathbf{S} = \left[ \mathbf{S}_1, \mathbf{S}_2, ..., \mathbf{S}_{d_m} \right] \in \mathbb{R}^{d_m \times d_r \times d_c}$, where $d_m$ and $d_n$ are spatial dimensions and $d_c$ is the number of feature maps. Then, we use the following convolutional projection to obtain $\mathbf{Q}, \mathbf{K} \in \mathbb{R}^{d_q}$ and $\mathbf{V} \in \mathbb{R}^{d_v}$ for the $j^\text{th}$ feature map of the next layer:
\begin{equation}
    \label{eq:attention_projection}
        \psi_j\left(\mathbf{S}, \mathbf{W}_j \right) : = \sum_{i=1}^{d_m} \text{conv2D} \left( \mathbf{S}_i, \mathbf{w}_{i, j} \right) + \mathbf{b}_j,
\end{equation}
where $\mathbf{w}_{i, j} \in \mathbb{R}^{f_s \times f_s}$ is the weight kernel. Next, the corresponding $\mathbf{Q}$, $\mathbf{K}$, and $\mathbf{V}$ are computed by,
\begin{equation}
\label{eq:rep}
    \begin{split}
        \mathbf{Q} = \sum_j \psi_j\left(\mathbf{S}, \mathbf{W}_j^q \right), \\
        \mathbf{K} = \sum_j \psi_j\left(\mathbf{S}, \mathbf{W}_j^k \right), \\
        \mathbf{V} = \sum_j \psi_j\left(\mathbf{S}, \mathbf{W}_j^v \right).
    \end{split}
\end{equation}
Then, the residual self-attention module is expressed as follows,
\begin{equation}
\label{eq:attention_final}
        \mathbf{S}_o = \alpha \cdot \text{Attention}\left( \mathbf{Q}, \mathbf{K}, \mathbf{V} \right) + \mathbf{S},
\end{equation}
where $\alpha$ is a learned scaling parameter, and the dot-product attention \cite{vaswani2017attention} is defined as,
\begin{equation}
    \label{eq:attention}
    \text{Attention}\left( \mathbf{Q}, \mathbf{K}, \mathbf{V} \right) = \text{SoftMax} \left( \frac{\mathbf{Q} \mathbf{K}^T}{\sqrt{d_q}} \mathbf{V} \right).
\end{equation}
In the generator network, incorporating residual self-attention and convolutional projection has several advantages. The attention values tend to grow, especially for the SoftMax edge locations, which results in substantially smaller gradient values. Residual connections limit this gradient vanishing scenario. Moreover, in several previous studies \cite{adalioglu2023saf, chen2021crossvit}, the representations of $\mathbf{Q}$, $\mathbf{K}$, $\mathbf{V}$ have been obtained using dense layers. This has contributed to computational complexity and exacerbated the gradient-vanishing problem. Therefore, we follow the 2-D convolutional projection for computing $\mathbf{Q}$, $\mathbf{K}$, $\mathbf{V}$ as depicted in \eqref{eq:attention_projection}.

The overall training loss of the ADA-Net approach for super-resolution is defined as,
\begin{equation}
\label{eq:objective}
    \begin{split}
    \mathcal{L}_{\text{ADA-Net}}\left( G, D, \mathbf{I}_l, \mathbf{I}_h \right) & = \mathcal{L}_A \left( G, D, \mathbf{I}_l, \mathbf{I}_h \right) \\
    & + \lambda \mathcal{L}_{\text{Spatial}}\left( G, \Phi, \mathbf{I}_l, \mathbf{I}_h \right) \\
    & + \beta \mathcal{L}_{\text{IDSpatial}}\left(G, \Phi, \mathbf{I}_h, \mathbf{I}_r \right) \\
    & + \gamma \mathcal{L}_{\text{Freq}}\left(G, \theta, \mathbf{I}_l \right) \\
    & + \vartheta \mathcal{L}_{\text{IDFreq}}\left(G, \theta, \mathbf{I}_h \right).
    \end{split}
\end{equation}
The adversarial loss component is the same with conventional Generative Adversarial Networks (GANs) \cite{goodfellow2020generative, ahishali2024r2c}, where the generator (G) and discriminator (D) models are sequentially updated through $\min_G \mathcal{L}_A\left( G, D, \mathbf{I}_l, \mathbf{I}_h \right)$ and $\max_{D} \mathcal{L}_A\left( G, D, \mathbf{I}_l, \mathbf{I}_h \right)$. The remaining parts of \eqref{eq:objective} are used for the generator training including $\mathcal{L}_{\text{Spatial}}$ and $\mathcal{L}_{\text{Freq}}$, i.e., spatial and frequency contrastive learning objectives with identity losses $\mathcal{L}_{\text{IDSpatial}}$ and $\mathcal{L}_{\text{IDFreq}}$.

The mutual information maximization \cite{cut, oord2018representation} is defined for a given query $\mathbf{I}_{l2h}$, which is the predicted high-resolution output from a positive-sample and low-resolution $\mathbf{I}_l$; and the negative sample group contains the rest of the images in $I_{L}$. These query, positive, and negative sample triplets are transformed to a joint subspace using $f: \mathbb{R}^{P \times N} \rightarrow \mathbb{R}^d$ to have, $\mathbf{f}_q, \mathbf{f}_p \in \mathbb{R}^d$, and $\mathbf{F}_n = \left[ \mathbf{f}_{n,1}, ..., \mathbf{f}_{n,K} \right] \in \mathbb{R}^{d \times K}$, respectively. We have used the following contrastive learning loss,
\begin{equation}
\label{eq:cont_loss}
    \mathcal{L}_{C} \left( \mathbf{f}_q, \mathbf{f}_p, \mathbf{F}_n \right) = - \log \left(  \frac{ e^{ \left( \mathbf{f}_q^T \mathbf{f}_p \right) / \tau} }{ e^{ \left( \mathbf{f}_q^T \mathbf{f}_p \right) / \tau} + \sum_{k=1}^K e^{\left( \mathbf{f}_q^T \mathbf{f}_{n, k} \right) / \tau} } \right),
\end{equation}
where $\mathcal{L}_{C}$ is integrated in $\mathcal{L}_{\text{Spatial}}$ and $\mathcal{L}_{\text{Freq}}$ of \eqref{eq:objective} which will be detailed next.

\subsection{Spatial Contrastive Learning}
Instead of directly using image samples, several layers of the generator $G$ are selected as feature extractors, and extracted features are given to $M$ different dense layers. Accordingly, $f: \mathbb{R}^{P \times N} \rightarrow \mathbb{R}^d$ stands for the combined mapping of the encoder of $G$ and multiple dense layers. Pixel-wise query sample features are obtained by feeding back the estimated high-resolution output into the generator to obtain $\mathbf{s}_q = \left[ \mathbf{s}_{q, 1}, ..., \mathbf{s}_{q, M} \right] \in \mathbb{R}^{d_m \times M}$, where $\mathbf{s}_{q, m} = f_m\left( G\left( \mathbf{I}_{l2h} \right) \right)$. Positive sample features, $\mathbf{s}_p = \left[ \mathbf{s}_{p, 1}, ..., \mathbf{s}_{p, M} \right] \in \mathbb{R}^{d_m \times M}$, are computed by $\mathbf{s}_{p, m} = f_m\left( G\left( \mathbf{I}_l \right) \right)$ using low-resolution images $\mathbf{I}_l$. Lastly, negative features, i.e., $\mathbf{S}_n = \left[ \mathbf{s}_{n, m}^k \right]_{m=1, k=1}^{M, \quad K} \in \mathbb{R}^{d_m \times M \times K}$, consist of different location pixel features obtained again by $G\left( \mathbf{I}_l \right)$. Given a set of $M$ number of Multi-layer Perceptrons (MLPs), $\Phi = \left\{ \phi_m \right\}^M_{m=1}$, contrastive loss for a triplet is defined:
\begin{equation}
\label{eq:cont_loss_pixel}
    \mathcal{L}_{\text{C}}^* = \frac{1}{M}\sum_{m=1}^M \mathcal{L}_{C} \left( \phi_m\left( \mathbf{s}_{q, m} \right), \phi_m\left( \mathbf{s}_{p, m} \right), \left[ \phi_m\left( \mathbf{s}_{n, m}^k \right) \right]_{k=1}^K \right),
\end{equation}

In this configuration, a total number of $N_s$ pixels are selected and features are extracted from the channel dimension of layers. Overall, final spatial contrastive learning loss is obtained by iterating triplets $\left\{ \mathbf{s}_q, \mathbf{s}_p, \mathbf{S}_n \right\}$ with $i \in \left\{1, 2, ..., N_s \right\}$ being the index operator,
\begin{equation}
\label{eq:cont_loss_total}
    \mathcal{L}_{\text{Spatial}} = \frac{1}{N_s} \sum_{i=1}^{N_s} \mathcal{L}_{\text{C}}^* \left( \Phi, \mathbf{s}_{q}^i, \mathbf{s}_{p}^i, \mathbf{S}_{n}^{N_s \setminus i} \right).
\end{equation}
Therefore, it follows that $K = N_s - 1$. Each MLP has three layers with $256$ neurons, i.e., $d_m = 256$, and we select $N_s = 256$ with $M = 5$. Their connections in the framework are shown in Fig. \ref{fig:framework}.

Identity spatial loss is expressed as $\mathcal{L}_{\text{IDSpatial}} = \frac{1}{N_s} \sum_{i=1}^{N_s} \mathcal{L}_{\text{C}}^* \left( \Phi, \Tilde{\mathbf{s}}_{q}^i, \Tilde{\mathbf{s}}_{p}^i, \Tilde{\mathbf{S}}_{n}^{N_s \setminus i} \right)$. The computation of $\Tilde{\mathbf{s}}_q$ is similar, but it is performed by $G\left( \Tilde{\mathbf{I}}_{h2h} \right)$ where $\Tilde{\mathbf{I}}_{h2h} = G\left( \mathbf{I}_h \right)$; $\Tilde{\mathbf{s}}_p$ and $\Tilde{\mathbf{S}}_n$ are extracted via $G\left( \mathbf{I}_{h} \right)$. The aim is to expose the generator network to a target domain, high-resolution image $\mathbf{I}_{h} \in I_H$, ensuring that if an already high-resolution image is given as input, the super-resolved output image undergoes limited alteration, therefore preserving the details. This property is especially useful for iterative super-resolution, where the generator is applied sequentially in inference.

\subsection{Patch-wise Frequency Domain Contrastive Learning}

Frequency domain representation of images reveals structural patterns that can be less obvious in the spatial domain. Hence, it is preferable to incorporate frequency domain information into GAN training \cite{jiang2021focal}. In this manner, patch-wise contrastive loss is obtained using frequency domain representation coefficients in the following:
\begin{equation}
\label{eq:cont_loss_patch}
    \mathcal{L}_{\text{freq}} = \frac{1}{N_f} \sum_{i=1}^{N_f} \mathcal{L}_{C} \left( \theta\left( \mathbf{z}_q^i \right), \theta\left( \mathbf{z}_p^i \right), \left[ \theta\left( \mathbf{z}_n^{k, N_f \setminus i} \right) \right]_{k=1}^{K_f} \right),
\end{equation}
where $\mathbf{z}_q^i, \mathbf{z}_p^i,  \mathbf{z}_n^{k, N_f \setminus i} \in \mathbb{C}^{d_f}$ are flattened patches extracted from location $i$ and $N_f \setminus i$ after applying 2D-FFT transformation shown in Fig. \ref{fig:framework}, and we can write $\mathbf{Z}_n^{N_f \setminus i} = \left[ \mathbf{z}_n^{1, N_f \setminus i}, \mathbf{z}_n^{2, N_f \setminus i}, ..., \mathbf{z}_n^{K_f, N_f \setminus i} \right]\in \mathbb{C}^{d_f \times K_f}$. Then, $\theta$ is a complex domain operating MLP which learns subspace mapping of $\theta: \mathbb{C}^{d_f \times d_f} \rightarrow \mathbb{R}^d$. It has three dense layers with $1024$ and $256$ neurons, respectively, in the hidden and output layers. We sample $N_f = 64$ number of patches with a size of $32 \times 32$, where $K_f = N_f - 1 = 63$ and $d_f = 256$.

Following the spatial identity loss, its frequency domain version is defined as $\mathcal{L}_{\text{IDFreq}} = \frac{1}{N_s} \sum_{i=1}^{N_s} \mathcal{L}_{C} \left( \theta\left( \Tilde{\mathbf{z}}_q^i \right), \theta\left( \Tilde{\mathbf{z}}_p^i \right), \left[ \theta\left( \Tilde{\mathbf{z}}_n^{k, N_f \setminus i} \right) \right]_{k=1}^{K_f} \right)$, where $\Tilde{\mathbf{z}}_q$ is extracted from $\Tilde{\mathbf{I}}_{h2h} = G \left( \mathbf{I}_h \right)$, $\Tilde{\mathbf{z}}_p$ and $\Tilde{\mathbf{Z}}_n$ are from $\mathbf{I}_h$.

Overall, contrastive learning improvements proposed by the ADA-Net approach in \cite{ahishali2025ada} can be summarized as follows: patch-wise frequency domain loss computation captures neighborhood correlation effectively, whereas relying only on pixel-wise sampling in spatial contrastive learning \cite{cut} can suffer from diminished contextual information during forward propagation. Moreover, integrating frequency domain representations improves the general structural preservation, as specific patterns are more prominently represented. Finally, extended regularization with combined identity losses reduces the network's sensitivity to minor changes in the input image.

\section{Experimental Evaluation}
\label{sec:experimental_evaluation}

\subsection{Experimental Setup}

For the evaluation of the segmentation task, we employ several well-known metrics derived from the computed global confusion matrix obtained by aggregrating the true and actual mask pixels, including true positives (TP), true negatives (TN), false positives (FP), and false negatives (FN). Specifically, Intersection over Union (IoU) is computed as,
\begin{equation}
    \text{IoU} = \text{TP} / (\text{TP} + \text{FP} + \text{FN}),
\end{equation}
and dice score, accuracy, and specificity are calculated as follows,
\begin{equation}
    \text{Dice Score} = \frac{2 \times \text{TP}}{(2 \times \text{TP} + \text{FP} + \text{FN})},
\end{equation}
\begin{equation}
    \text{Accuracy} = \frac{\text{TP} + \text{TN}}{\text{TP} + \text{TN} + \text{FP} + \text{FN}},
\end{equation}
\begin{equation}
    \text{Specificity} = \text{TN} / (\text{TN} + \text{FP}).
\end{equation}
Next, $F_2-\text{Score}$ is used with $a = 2$,
\begin{equation}
    F_a = (1 + a ^ 2) \frac{\text{Precision} \times \text{Sensitivity}}{a ^ 2 \times \text{Precision} + \text{Sensitivity}},
\end{equation}
where sensitivity and precision are defined in the following:
\begin{equation}
    \text{Sensitivity} = \text{TP} / (\text{TP} + \text{FN}),
\end{equation}
\begin{equation}
    \text{Precision} = \text{TP} / (\text{TP} + \text{FP}).
\end{equation}
Note the fact that we have adopted the $F_2-$Score for evaluation because the dice score already exhibits characteristics similar to $F_1-$Score as they both have a balanced weighting between precision and sensitivity. Moreover, when evaluating model performance, this study prioritizes specifically the detection of standing dead trees; therefore, the $F_2-$Score is more appropriate in the scope of this work as it has greater emphasis on sensitivity.

\subsection{Dataset Preparation}
\begin{figure}[]
    \centering
    \includegraphics[width=\linewidth]{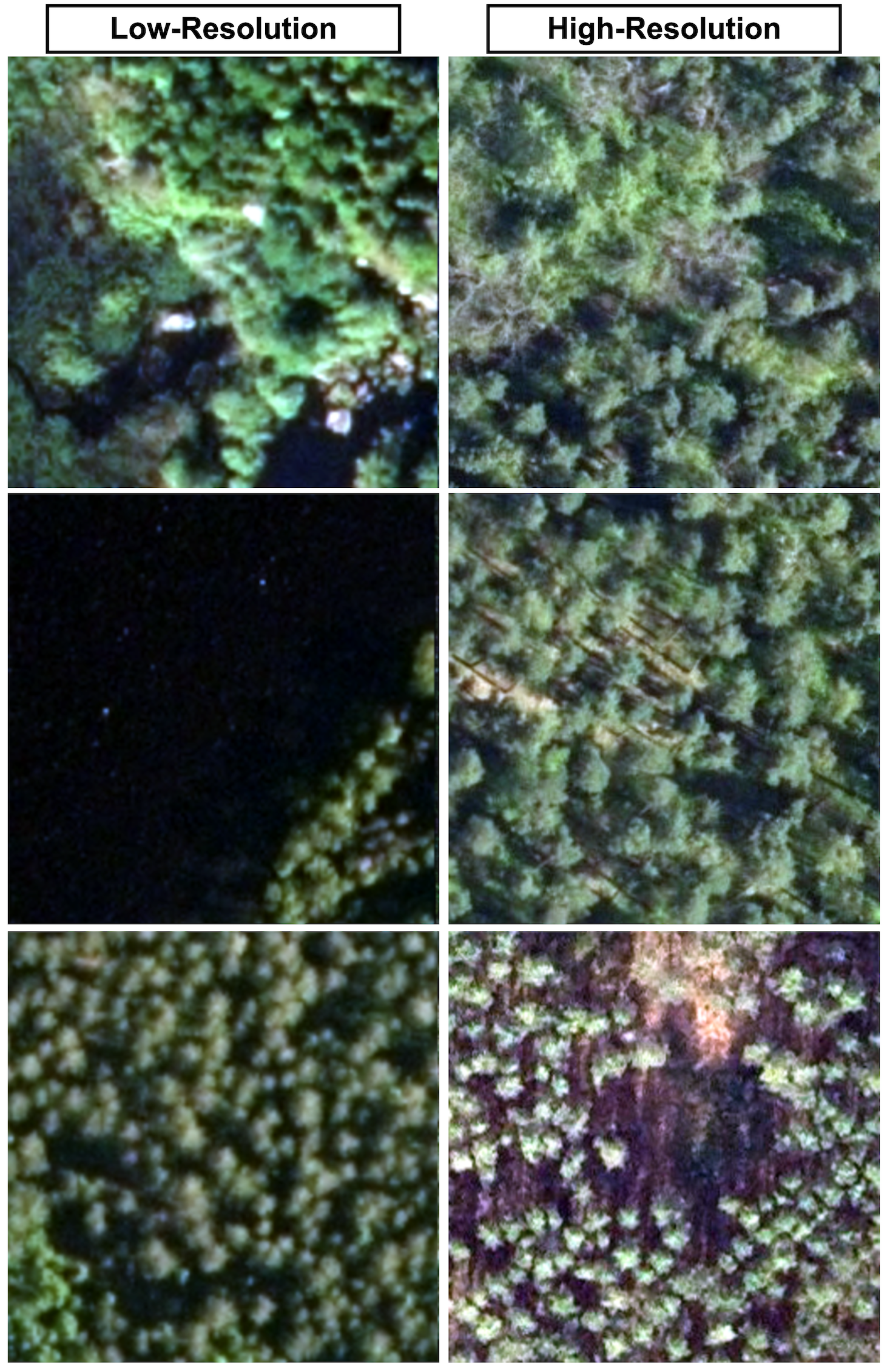}
    \caption{Several example image patches of low-resolution and high-resolution images, illustrated only RGB channels.}
\label{fig:dataset_samples}
\end{figure}

For the low-resolution images, we use originally $0.5$ m resolution multi-spectral aerial images from the National Land Survey of Finland \cite{finland1, finland2} between the years $2011$ and $2023$. The target domain high-resolution images have a resolution of $0.25$ m, and were acquired over Poland \cite{poland_orthoimagery} from $2021$ to $2022$. Both the source and target domain images have 4 bands consisting of R, G, B, and Near-Infrared (NIR) channels. In total, the dataset comprises $125$ low-resolution (Finland) images and $166$ high-resolution (Poland) image scenes with an average size of $6000 \times 6000$ pixels. These scenes are split into training, validation, and test sets using a 70:10:20 split. Furthermore, the training and validation scenes are divided into patches of $256 \times 256$ pixels, resulting in $35,142$ training and $6684$ validation patches covering both countries. Note that this is an unaligned dataset without corresponding low- and high-resolution image pairs designed; therefore, the dataset serves as an ideal benchmark for the blind super-resolution task. Sample image patches of this un-aligned dataset are provided in Fig. \ref{fig:dataset_samples}. Accordingly, low-resolution images exhibit overall low-quality, characterized by blurry appearances, poor contrast, and shadowed regions between the trees resulting from the image acquisition process. In contrast, high-resolution images are sharper and they maintain good contrast with details remaining visible even in darker regions. This mimics the real-world problem of super-resolution described in Section \ref{sec:introduction}: high-resolution sensors are typically more advanced and capable of producing high quality images, whereas low-resolution sensors are more cost‑effective with limited imaging performance.

\subsection{Model Training}
We use a comprehensive data augmentation pipeline for training. Spatial transformations are applied to both the images and label masks including random horizontal and vertical flips, and multiple 90-degree rotations. Moreover, several photometric augmentations are used, i.e., brightness and contrast adjustments, where brightness scaled by a factor uniformly sampled from $[0.8, 1.2]$ and contrast factor ranges smilarly from $0.8$ to $1.2$. Multiplicative noise is applied for each pixel by a random noise in $[0.9, 1.1]$. Finally, random gamma correction is performed by a uniformly sampled gamma from $[0.8, 1.2]$. For the performance evaluation, standing dead tree annotations of the high-resolution test split are considered. TreeMort-1T-UNet segmentation model is trained using its default hyper-parameter values proposed in \cite{rahman2025dual}. All Image-to-Image translation networks are trained for $60$ epochs, and hyperparameters are selected empirically: learning rate is $2\times10^{-6}$, $\lambda = \beta = \gamma = \vartheta = 0.5$, and $\tau = 0.07$ for the Ada-NET approach. All experimental evaluations have been carried out using PyTorch and we utilize the ADAM optimizer \cite{kingma2014adam} with its default decay rates defined as $\beta_1 = 0.9$ and $\beta_2 = 0.999$.

\subsection{Compared Methods}

We have compared different methods for blind super-resolution, including Contrastive Unpaired Translation (CUT) \cite{cut}, Fast-CUT \cite{cut}, and Cycle-Consistent Generative Adversarial Networks (Cycle-GANs) \cite{cycle_gan}. These models are trained using their default hyper-parameter values. The CUT method shares similar properties with ADA-Net approach; however it relies on spatial pixel sampling and does not incorporate extracted 2D-FFT features from image patches. Moreover, the CUT architecture is considerable deeper \cite{cut} consisting of more layers without self-attention blocks. Fast-CUT follows the same configuration as CUT, but it lacks the identity loss term during model training. As a result, its training is significantly faster because the model no longer requires target domain images as inputs. Finally, the Cycle-GAN framework has two generators and two discriminators. The forward generator learns the mapping from low-resolution to high-resolution image domain denoted as $G_F: I_L \rightarrow I_H$, whereas the inverse generator learns the reverse transformation, i.e., $G_I: I_H \rightarrow I_L$. Correspondingly, each discriminator aims to distinguish real images from generated ones within its respective domain.  A forward cycle is formed by passing the generated high-resolution image through the inverse generator yielding:
\begin{equation}
\Tilde{\mathbf{I}}_l = G_I \left( \mathbf{I}_{l2h}\right),   
\end{equation}
where $\mathbf{I}_{l2h} = G_F\left( \mathbf{I}_l\right)$. Similarly, the reverse cycle is defined as follows,
\begin{equation}
\Tilde{\mathbf{I}}_h = G_F \left( \mathbf{I}_{h2l}\right),
\end{equation}
where $\mathbf{I}_{h2l} = G_I\left( \mathbf{I}_h\right)$. Although it is generally more difficult to train Cycle-GANs, the introduced cycle-consistency loss $\mathcal{L}_{cyc} =
\|\mathbf{I}_l - \tilde{\mathbf{I}}_l\|_1 +
\|\mathbf{I}_h - \tilde{\mathbf{I}}_h\|_1$ computed after forward and backward cycles provides additional bidirectional reconstruction constraints that encourage meaningful domain mappings.

\subsection{Results}

\begin{table*}[]
\vspace{0.1cm}
\caption{The super-resolution performance is evaluated in terms of standing dead tree segmentation over the test high-resolution data. The segmentation TreeMort-1T-UNet model \cite{rahman2025dual} is trained on low-resolution (Baseline I) and super-resolved data produced by different super-resolution (SR) methods. The best and second-best results are highlighted in red and blue, respectively.}
\vspace{0.1cm}
\centering
\label{tab:results1}
\setlength{\tabcolsep}{10pt}
\renewcommand{\arraystretch}{1.4}
\resizebox{0.85\linewidth}{!}{
\begin{tabular}{@{}cccccccc@{}}
\toprule
\textbf{SR Method} & \textbf{Dice Score} & \textbf{F2-Score} & \textbf{IoU} & \textbf{Accuracy} & \textbf{Precision} & \textbf{Specificity} & \textbf{Sensitivity} \\ \midrule
\textbf{Baseline I} & $0.3869$ & $0.3046$ & $0.2398$ & \textcolor{blue}{\boldmath$0.9963$} &  \textcolor{red}{\boldmath$0.7037$} &  \textcolor{red}{\boldmath$0.9995$} & $0.2667$ \\
\textbf{Cycle-GAN} & $0.4452$ & \textcolor{blue}{\boldmath$0.5014$} & $0.2864$ & $0.9940$ & $0.3752$ & $0.9960$ &  \textcolor{red}{\boldmath$0.5474$} \\
\textbf{Fast-CUT} &  \textcolor{blue}{\boldmath$0.5048$} & $0.4392$ & \textcolor{blue}{\boldmath$0.3376$} &  \textcolor{red}{\boldmath$0.9965$} & \textcolor{blue}{\boldmath$0.6722$} & \textcolor{blue}{\boldmath$0.9991$} & $0.4042$ \\
\textbf{CUT} & $0.4934$ & $0.4857$ & $0.3275$ & $0.9956$ & $0.5067$ & $0.9979$ & $0.4808$ \\
\textbf{ADA-Net} &  \textcolor{red}{\boldmath$0.5406$} &  \textcolor{red}{\boldmath$0.5383$} &  \textcolor{red}{\boldmath$0.3704$} & $0.9960$ & $0.5445$ & $0.9980$ & \textcolor{blue}{\boldmath$0.5367$} \\ \bottomrule
\end{tabular}}
\end{table*}

%
%
%
%
We present two complementary sets of results addressing key questions regarding both super-resolution and the domain shift challenges related to global-scale standing dead tree mapping. First, we evaluate whether applying super-resolution techniques can adapt a standing dead tree segmentation model to a new high-resolution domain without requiring any annotations from the target domain. Secondly, we show that applying super-resolution techniques improves the standing dead tree segmentation model performance when only a limited number of annotations from the new domain are available.

\begin{figure*}[]
    \centering
    \includegraphics[width=\linewidth]{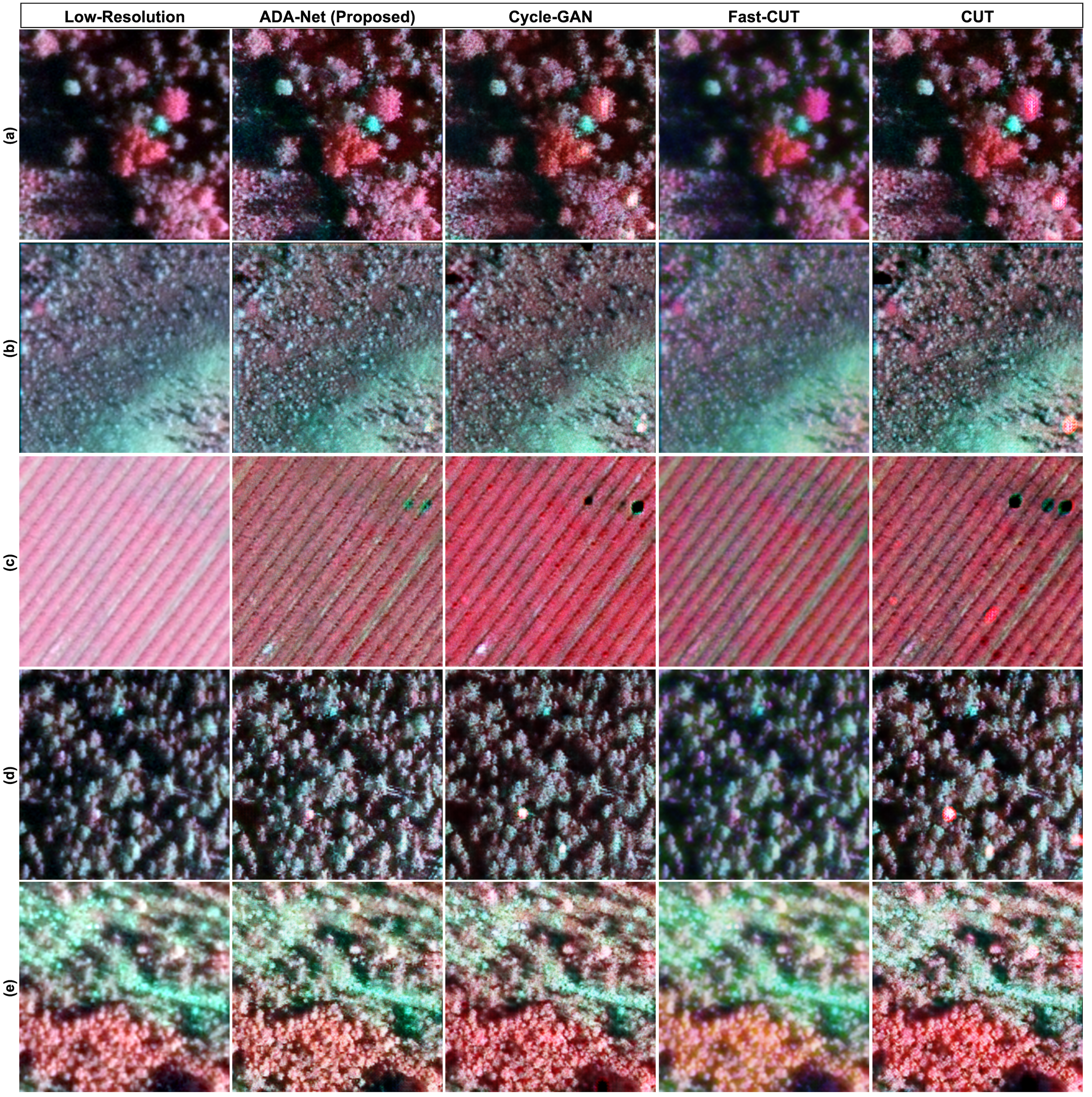}
    \caption{False-color representations of extracted test image patches (a-e) with $256 \times 256$ dimensions are formed by assigning NIR, R, and G channels to RGB, respectively. The corresponding super-resolved images are produced using highlighted methods including ADA-Net, Cycle-GAN, Fast-CUT, and CUT.}
\label{fig:visual_results1}
\end{figure*}

\begin{figure*}[]
    \centering
    \includegraphics[width=0.92\linewidth]{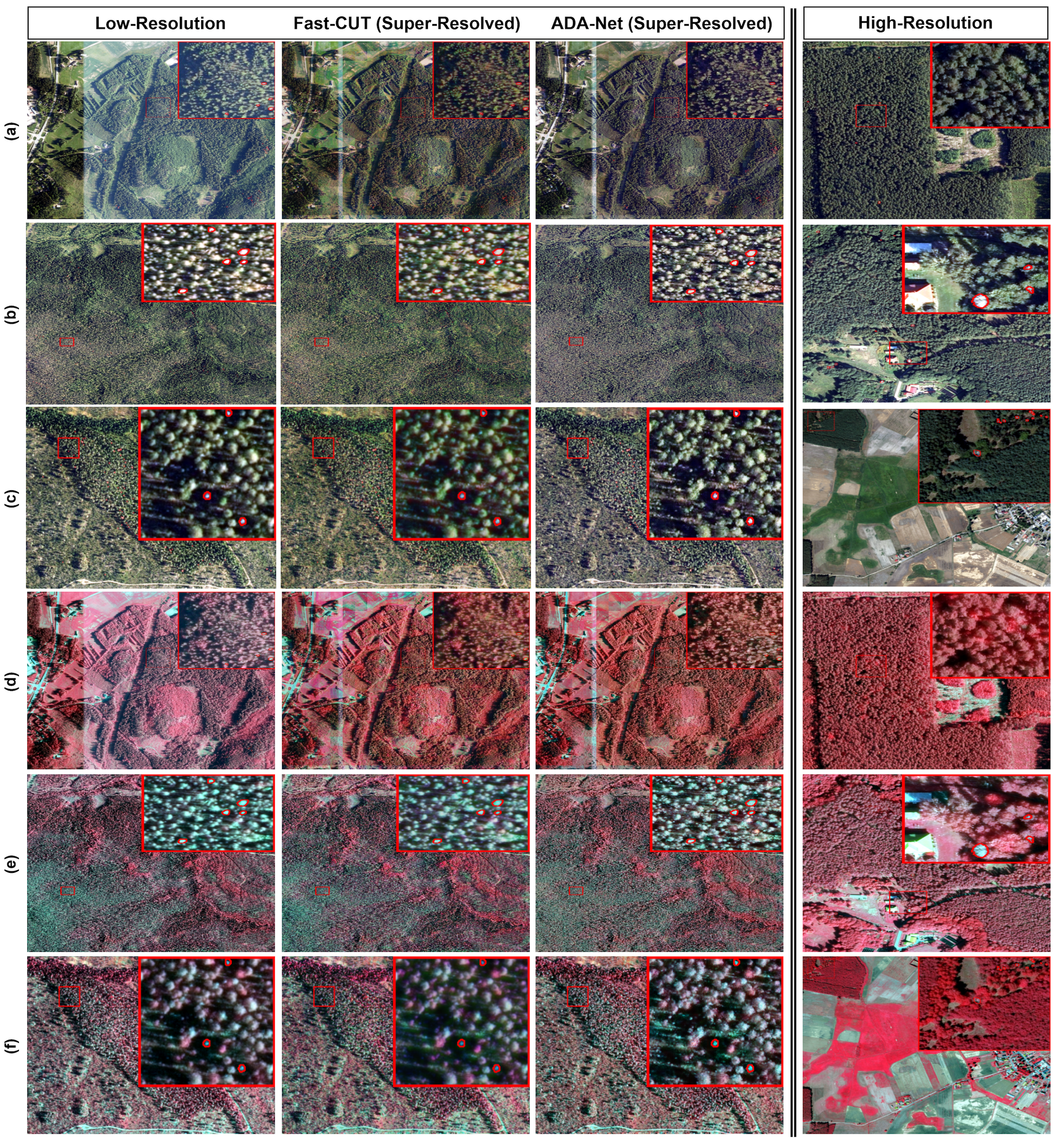}
    \caption{Examples of full-scene low-resolution images and their super-resolved versions are given for both RGB (a–c) and false-color representations (d–f), where NIR, R, and G channels are mapped to RGB. The last column demonstrates samples from the high-resolution reference dataset. Standing dead trees are displayed with red polygon annotations. Rectangles represent enlarged regions to facilitate detailed visual inspection.}
\label{fig:visual_results2}
\end{figure*}

When TreeMort-1T-UNet segmentation model is trained using both low-resolution data (Baseline I) and super-resolved versions, its performance on high-resolution test data is presented in Table \ref{tab:results1}. Accordingly, the images super-resolved by ADA-Net have achieved the best overall performances across most metrics, except for precision. For accuracy and specificity, all methods can achieve comparable levels. Notably, compared to segmentation results using low-resolution (Baseline), the Dice score improves by more than $0.15$ using ADA-Net to $0.54$. The second set of experiments, presented in Table \ref{tab:results2}, further investigates the effect of combining high-resolution data with the generated high-resolution (super-resolved) data by the ADA-Net approach. This hybrid usage of real and synthetic high-resolution data has achieved the highest segmentation performance, obtaining a Dice score of $63.86\%$ and approximately $70\%$ in $F_2-$Score outperforming Baseline II, where the model is trained only with high-resolution data. Overall, these results demonstrate that domain-specific features important for discriminating standing dead trees are successfully preserved during the super-resolution process.

\begin{table}[h]
\centering
\vspace{0.1cm}
\caption{The TreeMort-1T-UNet \cite{rahman2025dual} segmentation model is trained using high-resolution (HR) data (Baseline II), and the combination of HR and super-resolved (HR + SR) data by ADA-Net, test segmentation results are presented over the HR data.}
\vspace{0.1cm}
\label{tab:results2}
\setlength{\tabcolsep}{12pt}
\renewcommand{\arraystretch}{1.4}
\resizebox{0.8\linewidth}{!}{
\begin{tabular}{@{}ccc@{}}
\toprule
            & \makecell{\textbf{HR Data} \\ \textbf{(Baseline II)}}        &  \hspace{0.2cm} \makecell{\textbf{HR + SR} \\ \textbf{Data}} \hspace{0.2cm} \\ \midrule
\textbf{Dice Score}           & $0.6144$ & \boldmath$0.6386$ \\
\boldmath$F_2-$\textbf{Score} & $0.6674$ & \boldmath$0.6957$ \\
\textbf{IoU}                  & $0.4434$ & \boldmath$0.4691$ \\
\textbf{Accuracy}             & $0.9958$ & \boldmath$0.9961$ \\
\textbf{Precision}            & $0.5427$ & \boldmath$0.5619$ \\
\textbf{Specificity}          & $0.9972$ & \boldmath$0.9973$ \\
\textbf{Sensitivity}          & $0.7080$ & \boldmath$0.7397$ \\ \bottomrule
\end{tabular}}
\vspace{0.2cm}
\end{table}

For qualitative evaluation, we present example super-resolved images by the proposed approach and competing benchmark methods in Fig. \ref{fig:visual_results1}. Notable, generated images by Fast-CUT contains the least hallucination; in particular, Cycle-GAN and CUT methods substantially hallucinate in specific regions of \ref{fig:visual_results1}c, whereas ADA-Net results in only slight hallucination. This observation may explain why Fast-CUT achieves the best segmentation performance among the competing methods as given in Table \ref{tab:results1}. On the other hand, despite this advantage, Fast-CUT fails to fully super-resolve the majority of images from a qualitative perspective as presented in Fig. \ref{fig:visual_results1}, unlike the proposed approach. Both Cycle-GAN and CUT models have introduced visible artifacts especially in Fig. \ref{fig:visual_results1}b - \ref{fig:visual_results1}d characterized by red dashed-like spots with CUT showing more severe artifacts. Color-shifts are also noticeable across all competing methods, whereas the proposed approach is able to preserve color consistency. Overall, ADA-Net approach has demonstrated the best performance compared to other methods based on visual inspection of the generated images.

Further example images are illustrated in Fig. \ref{fig:visual_results2} alongside their super-resolved versions produced by Fast-CUT and ADA-Net models, selected based on the quantitative evaluation in Table \ref{tab:results1}. These examples correspond to full-scene images with a large number of pixels, where each frame is reconstructed by aggregating the outputs of patch-wise processing with a patch size of $256 \times 256$. While the image details of low-resolution images are enhanced, the super-resolution approach also functions as a general image restorer in RGB domain, e.g., saturation correction is visible in Fig. \ref{fig:visual_results2}a. Visual inspection of both RGB and false-color illustrations reveals that the ADA-Net approach has achieved superior image details, closely resembling the high-resolution image samples from the target reference domain. Notably, in addition to generated images shown in false-color images, RGB domain representations produced by ADA-Net demonstrates superior preservation of color fidelity as seen in Fig. \ref{fig:visual_results2}c, whereas Fast-CUT alters the color appearance relative to the original low-resolution image.

\subsection{Limitations and Discussion}

Real-world image super-resolution differs fundamentally from conventional studies \cite{lu2022transformer, yang2014single, xiao2024frequency, 11329206, 11435474, cai2019toward}  where low-resolution images are typically obtained by downsampling high-resolution images. As previously discussed in Section \ref{sec:introduction}, this discrepancy is due to the fact that low-resolution images are often acquired by lower quality sensors, leading to complex and unknown degradations. Therefore, training paired super-resolution models becomes challenging due to the lack of accurately aligned ground-truth data. In this study, we address this challenge by formulating the super-resolution problem as an Image-to-Image translation task, where it estimates without explicitly modeling intermediate upsampling operators. The learned transformation implicitly captures complex and potentially non-linear relations between low-resolution and high-resolution image distributions that are otherwise difficult to express using classical super-resolution or handcrafted priors. 

While our results demonstrate potential performance improvements when the restored images are evaluated for the downstream task of standing dead tree segmentation, performance comparisons are obviously limited to other Image-to-Image translation approaches. This is because because most existing studies rely on strong assumptions such as the availability of paired low and high-resolution images and/or prior knowledge of the degradation model and resolution quality. On the other hand, we aim to draw more attention to real-world super-resolution scenarios in which the task is performed in fully blind manner without any explicit assumptions about the relationship between the source and target domains. In this way, we believe this direction is important for enabling more realistic benchmarking and wider model comparisons in future studies.

Formulating super-resolution as Image-to-Image translation problem inherently necessitates the use of \textit{generative models} as traditional regression based models need paired training data. A known drawback of generative models is their tendency to \textit{hallucinate} image content. This phenomenon has been more evident for some models in certain cases, for example, in Fig. \ref{fig:visual_results1}c-d, where healthy trees can be transformed to dead trees or vice versa after the super-resolution. We highlight a soft trade-off between preserving the image semantic consistency and reconstructing fine-grained details while acting as a generic image restoration model. Nevertheless, the ADA-Net model has the same inherited hallucination-related properties commonly observed in GAN-based approaches \cite{9035107, Banerjee_2020_WACV}. It is observed by \cite{9035107} that many popular GAN models have common artifacts, and the authors demonstrate that it is possible to develop a GAN simulator to train a real/fake image classifier which can successfully detect images generated by recent GAN models. Secondly, the study in \cite{Banerjee_2020_WACV} argues that eliminating GAN hallucinations is not even fully achievable; however, the authors propose a technique for generating realistic images despite this limitation. Overall, due to the lack of corresponding true high-resolution images and possible existing hallucinations, it is not possible to report conventional image quality metric measured between the restored and true high-quality images such as Peak Signal-to-Noise Ratio (PSNR) or Structural Similarity Index (SSIM) metrics. Therefore, we argue that blind real-world super-resolution approach needs to be evaluated through prioritizing task-driven or goal-oriented metrics. For instance, by utilizing available dead tree annotations after super-resolution, we focus on preserving dead tree class information in this study, thereby we are able to present indirect but feasible measure of network hallucination with respect to the selected target application.

\section{Conclusion}
\label{sec:conclusion}

In this study, we present a super-resolution approach suitable for real-world scenarios, where the low-resolution image domain is not represented by synthetically downsampled high-resolution images. Instead, the proposed approach learns to map from a general low-resolution domain to a high-resolution image domain using unpaired training samples. This is a significant advantage considering the fact that no corresponding low-resolution and high-resolution image pairs available in the real-world. Furthermore, since low-resolution images are often acquired using inexpensive, low-quality sensors, the proposed framework effectively functions as a generic image restoration model, substantially enhancing the overall visual quality of the images. We have evaluated the super-resolution performance in the context of segmenting standing dead trees. Accordingly, the ADA-Net approach has achieved superior segmentation performance compared to other image-to-image translation methods and against multiple baselines. These results demonstrate that learned super-resolution transformation preserve semantically meaningful features critical for downstream analysis. Overall, this work introduces a pioneering application for blind super-resolution without prior assumptions about the type or intensity of image degradation. The proposed framework exhibits strong architectural generality, the same framework can be utilized for different scaling factors and image domains with minimal task-specific modifications. In future work, we plan to investigate other applications in forest health monitoring and remote sensing, potentially utilizing hyperspectral, LiDAR and multimodal datasets to enhance both reconstruction fidelity and semantic understanding. This work will promote scalable, low-cost environmental monitoring pipelines while reducing dependency on expensive high-resolution sensors.

\section*{Acknowledgments}
The work has been funded by the European Union (ERC-2023-STG grant agreement no. 101116404) and the Research Council of Finland within the MULTIRISK project (grant no. 353262, NextGenerationEU funding). Views and opinions expressed are, however, those of the author(s) only and do not necessarily reflect those of the European Union or the European Research Council Executive Agency. Neither the European Union nor the granting authority can be held responsible for them. The authors greatly acknowledge computational resources provided by CSC – IT Center for Science, Finland.

\bibliographystyle{IEEEtran}
\bibliography{IEEEtran}

\vfill

\end{document}